\documentclass[10pt,twocolumn,letterpaper]{article}

\usepackage{wacv}
\usepackage{times}
\usepackage{epsfig}
\usepackage{graphicx}
\usepackage{amsmath}
\usepackage{amssymb}
\usepackage{booktabs}
\usepackage{subeqnarray}
\usepackage{algorithm}
\usepackage{algorithmic}
\usepackage{colortbl}
\usepackage{enumitem}
\usepackage{multirow}
\usepackage{subcaption}

\usepackage{amsmath,amsfonts,bm}

\newcommand{\taski}{{\mathcal{T}_i}}

\def\eqref#1{equation~\ref{#1}}

\def\1{\bm{1}}

\DeclareMathAlphabet{\mathsfit}{\encodingdefault}{\sfdefault}{m}{sl}
\SetMathAlphabet{\mathsfit}{bold}{\encodingdefault}{\sfdefault}{bx}{n}

\newcommand{\task}{\mathcal{T}}

\def\wacvPaperID{308} 

\wacvfinalcopy 

\ifwacvfinal
\usepackage[breaklinks=true,bookmarks=false]{hyperref}
\usepackage[capitalize]{cleveref}
\crefname{section}{Sec.}{Secs.}
\Crefname{section}{Section}{Sections}
\Crefname{table}{Table}{Tables}
\crefname{table}{Tab.}{Tabs.}
\else
\usepackage[pagebackref=true,breaklinks=true,colorlinks,bookmarks=false]{hyperref}
\usepackage[capitalize]{cleveref}
\crefname{section}{Sec.}{Secs.}
\Crefname{section}{Section}{Sections}
\Crefname{table}{Table}{Tables}
\crefname{table}{Tab.}{Tabs.}

\fi

\begin{document}

\title{Few-Shot Learning of Compact Models via Task-Specific Meta Distillation}


\author{
	Yong Wu$^{1}$,
	Shekhor Chanda$^{2}$, 
	Mehrdad Hosseinzadeh$^3$,
	Zhi Liu$^{1*}$, 
	Yang Wang$^{4}$\thanks{Corresponding authors: Zhi Liu and Yang Wang.}\\
	\\
	$^1$Shanghai University,\quad  $^2$University of Manitoba, \quad $^3$ Huawei Technologies Canada,\quad $^4$Concordia University\\
	{\tt\small yong\_wu@shu.edu.cn}, \quad {\tt\small chandas@myumanitoba.ca}, \quad {\tt\small mehrdad.hosseinzadeh@live.com}\\
	{\tt\small liuzhi@staff.shu.edu.cn}, \quad 	{\tt\small yang.wang@concordia.ca}			
}

\maketitle
\begin{abstract}
We consider a new problem of few-shot learning of compact models. Meta-learning is a popular approach for few-shot learning. Previous work in meta-learning typically assumes that the model 
architecture during meta-training is the same as the model architecture used 
for final deployment. In this paper, we challenge this basic assumption. For 
final deployment, we often need the model to be small. But small models usually 
do not have enough capacity to effectively adapt to new tasks. In the mean 
time, we often have access to the large dataset and extensive computing power 
during meta-training since meta-training is typically performed on a server. In 
this paper, we propose \emph{task-specific meta distillation} that simultaneously learns two models in 
meta-learning: a large teacher model and a small student model. These two 
models are jointly learned during meta-training. Given a new task during 
meta-testing, the teacher model is first adapted to this task, then the 
adapted teacher model is used to guide the adaptation of the student model. The 
adapted student model is used for final deployment. We demonstrate the 
effectiveness of our approach in few-shot image classification using 
model-agnostic meta-learning (MAML). Our proposed method outperforms other 
alternatives on several benchmark datasets.
\end{abstract}

\section{Introduction}
\label{sec:introduction}

\begin{figure}[ht]
\begin{center}
\includegraphics[width= 0.45\textwidth]{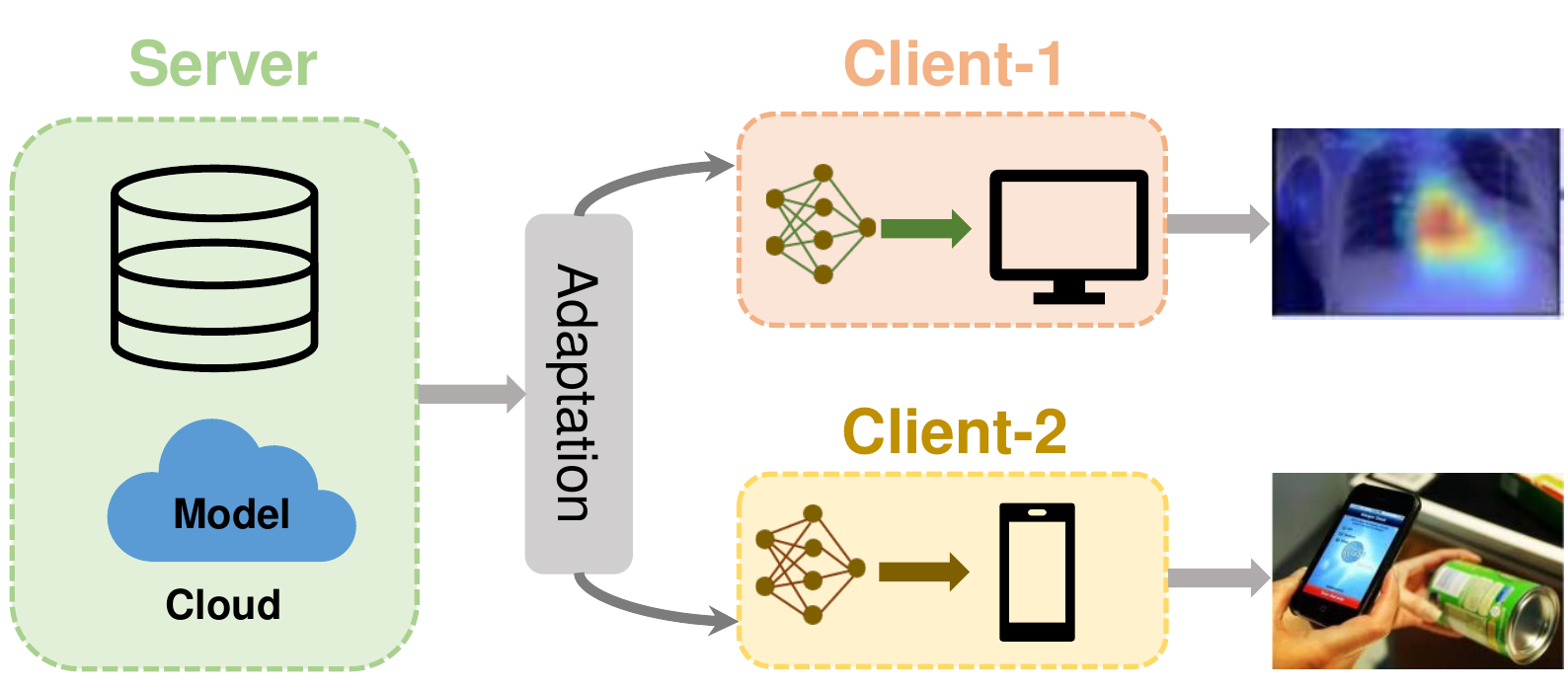}
\end{center}
   \caption{{\bf Motivation of our problem}. Consider the image classification 
   problem with a server (\eg cloud vendor) and many clients. The cloud vendor 
   has extensive computing power and training images of many classes. Each 
   client may want to solve a client-specific image classification problem with 
   some new classes not covered by the cloud vendor. For example, one client 
   wants to classify different medical images, while another client wants to 
   classify various merchandise in a particular store. Each client only has 
   few-shot data. The cloud vendor can train some model on the server side. Due 
   to privacy concerns, clients do not want to upload their own data to the 
   cloud vendor. Instead, each client adapts the model from the cloud vendor to 
   his/her specific image classification task on the client side. We would like 
   the final model for deployment on the client side to be small.}
\label{fig:problem}
\end{figure}

\begin{figure}[ht]
\begin{center}
\includegraphics[width= 0.4\textwidth]{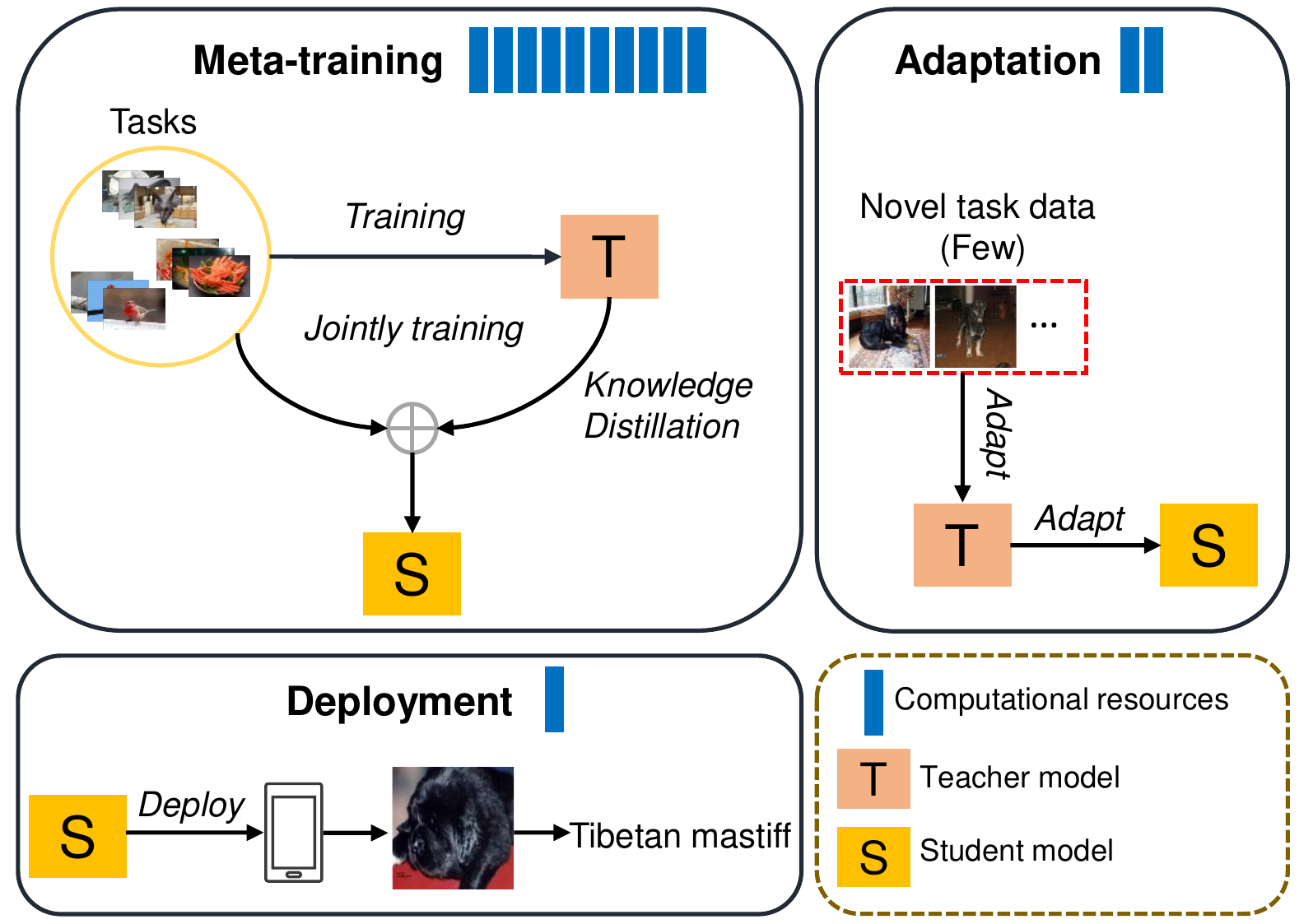}
\end{center}
\caption{{\bf Key idea of our approach}. On the server side, we jointly learn a 
large teacher model and a small student model in a meta-training framework. At 
the client side, the client first performs an adaptation stage. During this 
stage, the teacher model is first adapted to the task, then the adapted teacher 
model is used to guide the adaptation of the student model via distillation. 
The adapted student model is then used for final deployment. Different stages 
(meta-training, adaptation, deployment) of this pipeline involve different 
levels of computational resources.}
\label{fig:motivation}
\end{figure}

Meta-learning techniques, such as model-agnostic meta-learning 
(MAML)~\cite{2017_finn_MAML}, have been successfully applied in many computer 
vision and machine learning tasks, such as few-shot 
learning~\cite{2017_finn_MAML}, domain adaptation~\cite{li20_eccv}, domain 
generalization~\cite{li18_aaai}, etc. Meta-learning consists of a meta-training 
stage and a meta-testing stage. During meta-training, a global model is learned 
from a set of tasks. For example, in the case of few-shot learning (FSL), each 
task is a few-shot classification problem. During meta-testing, the learned 
global model can be adapted to a new few-shot classification problem with only 
few labeled examples. Meta-training is typically done on a central server 
where it is reasonable to assume access to extensive computational resources. 
In contrast, meta-testing is presumably done by an end-user or client who may 
not have the same computational resources as the central server, especially if 
the application of the client needs to run on low-powered edge devices. 
Existing meta-learning literature has largely overlooked this gap of 
computational resources. Existing meta-learning~(or more broadly, few-shot 
learning in general) approaches typically assume that the architecture of the 
model during meta-training is the same as the one used by the client for the 
final deployment. In this paper, we challenge this basic assumption of existing 
meta-learning solutions. We propose a new problem setting that takes into 
account of different levels of available computational resources during meta-training and meta-testing. 

Our problem setting is motivated by a practical scenario (shown in 
Fig.~\ref{fig:problem}) consisting of a server and many clients. For example, 
the server can be a cloud vendor that provides pretrained image classification 
models (possibly via web API). On the server side, the cloud vendor may have a 
large image dataset with many object classes. The cloud vendor typically has 
access to significant computational resources to train very large models. We 
also have clients who are interested in solving some application-specific image 
classification problems. Each client may only be interested in recognizing a 
handful of object classes that are potentially not covered by the training 
dataset from the server side. For example, one client might be a medical doctor 
interested in recognizing different tumors in medical images, while another 
client might be a retail owner interested in classifying different merchandise 
in a store. Because of the cost of acquiring labeled images, each client may 
only have a small number of labeled examples for the target application. Due to 
privacy concerns, clients may not want to send their data to the cloud vendor. 
In this case, a natural solution is for a client to re-use a pretrained model 
provided by the cloud vendor and perform few-shot learning to adapt the 
pretrained model to the new object classes for the target application.

At first glance, the scenario in Fig.~\ref{fig:problem} is a classic 
meta-learning problem. The cloud vendor can perform meta-training on the server 
to obtain a global model. On the client side, the client performs two steps. 
The first step (called \emph{adaptation}) is to adapt the global model from the 
server side to the target application. For example, the adaptation step in MAML 
performs a few gradient updates on the few-shot data. After the adaptation, the 
second step (called \emph{deployment}) is to deploy the adapted model for the 
end application. The combination of these two steps (adaptation and deployment) 
is commonly known as ``meta-testing'' in meta-learning. In this paper, we make 
a distinction between adaptation and deployment since this distinction is 
important for motivating our problem setting. Our key observation is that the 
available computing resources are vastly different in these different stages. 
The meta-training stage is done on a server or the cloud with significant 
computing resources. The adaptation stage is often done on a client's local 
machine with moderate computing powers (e.g., desktop or laptop). For the deployment, we may only have access to very limited computing power if the model is deployed on an edge 
device. If we want to use classic meta-learning in this case, we have to 
choose a small model architecture to make sure that the final model can be 
deployed on the edge device. Unfortunately, previous 
work~\cite{li20_aaai,mirzadeh20_aaai} has shown that a small model may not have 
enough capacity to consume the information from a large amount of data 
available during meta-training, so the learned model may not effectively adapt 
to a new task.

In this paper, we propose a new approach called \emph{task-specific meta distillation} to solve this problem. The key idea of our approach is 
illustrated in Fig.~\ref{fig:motivation}. During meta-training, we 
simultaneously learn a large teacher model and a small student model. Since the 
teacher model has a larger capacity, it can better adapt to a new task. During 
meta-training, these two models are jointly learned in a way that the teacher 
model can effectively guide the adaptation of the student model. During the 
adaptation step of meta-testing, we first adapt the teacher model to the target 
task, then use the adapted teacher to guide the adaptation of the student model 
via knowledge distillation. Finally, the \emph{adapted} student model is used 
for the final \emph{deployment}. In this paper, we apply our proposed approach 
to improve few-shot image classification with MAML. But our technique is 
generally applicable in other meta-learning tasks beyond few-shot learning.

The contributions of this work are manifold. First, previous work in 
meta-learning has largely overlooked the issue of the computational resource 
gap at different stages of meta-learning. This issue poses challenges in the 
real-world adoption of meta-learning applications. In this paper, we consider the problem of few-shot learning of compact models to
address this gap. Second, we propose a meta-learning approach that jointly 
learns a large teacher model and a small student model. During meta-testing, 
the adapted teacher model is used to distill task-specific knowledge for 
adapting the student model. The adapted student is used for final deployment. 
Finally, we apply the proposed approach to improve MAML-based few-shot image 
classification. Our proposed method significantly outperforms vanilla MAML. 
Although we focus on few-shot learning with MAML in this paper, the proposed 
method is generally applicable to other meta-learning tasks.






\section{Related Work}
In this section, we review several lines of research relevant to our work.

\noindent{\bf Knowledge distillation.} Knowledge Distillation 
(KD)~\cite{2019_Park_rkd,romero15_iclr,2017_kd_attention,2019_ABdistill,2019_ccdistillation,
2019_tian_crd,2018_kdsvd,2019_vid_CVPR,2019_similarity_ICCV,
2021_pkt_tnnls,2017_nts,2014_li_learning,Liu_2019_CVPR}
 is a widely used technique for model compression. KD aims to 
transfer the knowledge from a large model (called \emph{teacher}) to a small 
model (called \emph{student}). Most KD methods optimize a loss function that 
captures some form of dis-similarity between the teacher model and the student 
model, so that the student model is learned to mimic the teacher model. Hinton 
et~al.,~\cite{2015_hinton_distilling} introduce a KD method by defining a 
divergence loss between the soft outputs of the teacher model and the student 
model. Li et~al.,~\cite{2014_li_learning} propose a similar approach. In 
addition 
to soft outputs, there have been KD methods based on various other information, 
such as intermediate-layer features~\cite{romero15_iclr}, attention 
maps~\cite{2017_kd_attention}, hidden neuron activations~\cite{2019_ABdistill}, 
relationship of 
samples~\cite{2019_Park_rkd,2019_ccdistillation,2019_similarity_ICCV}, mutual 
information~\cite{2019_vid_CVPR}, correlation and higher-order 
dependencies~\cite{2019_tian_crd}, etc.

    

    
    
    

    
    
    


\noindent{\bf Few-shot learning and meta-learning.} The goal of few-shot 
learning (FSL) is to quickly learn new concepts from only a small number of 
training data. For example, in few-shot image classification, we are presented 
with some new classes that never appear during training. Our goal is to learn a 
model that can recognize those new classes when we only have very few training 
examples for each class. 
Meta-learning~\cite{2017_protonet,2017_finn_MAML,2018_reation_net,
2016_matchingnet,chen2019closerfewshot,Lee_2019_CVPR,Chi_2022_CVPR,Jiang_2022_CVPR,VS_2022_WACV,baik2021meta}
 has been a popular approach for few-shot image classification. In 
meta-learning, the model is learned from a set of \emph{tasks} during training, 
where each task is a few-shot image classification problem. For a new few-shot 
image classification task, the model is adapted to this new task using few-shot 
training examples of the task. Model-agnostic 
meta-learning~(MAML)~\cite{2017_finn_MAML} learns a good initialization of the 
model, so that the model can adapt to a new few-shot classification task with 
only a few gradient updates. Prototypical Network~\cite{2017_protonet} learns a 
metric space so that classification can be achieved by computing distances to 
class prototypes. Matching Network~\cite{2016_matchingnet} learns a model to 
map a small labeled support set and an unlabeled example to its label. Relation 
network~\cite{2018_reation_net} uses a similar idea. 
MetaOptNet~\cite{Lee_2019_CVPR} learns representations with a discriminatively 
trained linear predictor. MeTAL~\cite{baik2021meta} learns to adapt to various tasks via a loss function instead of hand-designing an auxiliary loss.

\noindent{\bf Meta-learning for knowledge distillation.} Recently, there has 
been 
work~\cite{2020_metadistil,pan21_acl,zhang2020knowledge,jang19_icml,li20_cvpr,bai20_aaai,liu20_eccv,pan21_acl,shen21_aaai,LIM2021327,liu2021few,zhong2022metadmoe}
 on combining knowledge distillation and meta-learning. Most of these works 
focus on using meta-learning ideas to improve knowledge distillation. 
MetaDistill~\cite{2020_metadistil} uses meta-learning to learn a better teacher 
model that is more effective to transfer knowledge to a student model. Instead 
of fixing the teacher, the method uses the feedback from the student model to 
improve the teacher model. Meta-KD~\cite{pan21_acl} uses a meta-teacher model 
to capture transferable knowledge across domains, then uses it to transfer 
knowledge to students. MetaDistille~\cite{liu20_eccv} generates better soft 
targets by using a label generator to fuse the feature maps. It uses 
meta-learning to optimize the label generator. Jang~et~al.,~\cite{jang19_icml} 
use meta-learning to learn what knowledge to transfer from the teacher model to 
the student model. Meta-DMoE~\cite{zhong2022metadmoe} incorporates Mixture-of-Experts (MoE) as teachers to address multiple-source domains shift. The above-mentioned works improve knowledge distillation 
using meta-learning, but they do not consider few-shot learning. There are some 
recent works~\cite{bai20_aaai,li20_cvpr,shen21_aaai} on learning the student 
model with few-shot examples using KD. But they only consider using few-shot 
examples of \emph{known} classes for learning the student model. They do not 
consider the few-shot learning of new classes. Lim~et~al.,~\cite{LIM2021327} propose to improve Efficient-PrototypicalNet performance with self knowledge distillation. Different from KD, the teacher and student models for self knowledge distillation have same network structure. Liu and Wang~\cite{liu2021few} propose a model that learns representation through online self-distillation. They introduce a special data augmentation-CutMix~\cite{yun2019cutmix} to improve few-shot learning performance.

To the best of our knowledge, ours is the first work on using KD  with 
meta-learning for few-shot learning.











\section{Preliminary and Background}
\label{sec:background}
In this section, we briefly introduce some background knowledge and terminology relevant to our work.

\noindent{\bf Knowledge distillation.} The goal of knowledge distillation is to 
transfer the knowledge of a large teacher model $f_{\psi}$ parameterized by 
$\psi$ when training a small student model $g_{\theta}$ parameterized by 
$\theta$. Given a labeled dataset $D$ and a pretrained teacher model 
$f_{\psi}$, we can learn the student model $g_{\theta}$ by optimizing the 
following loss function:
\begin{equation}
\label{eq:loss}
\min_{\theta}\mathcal{L}_{S}(\theta; D)+\mathcal{L}_{\mathit{KD}}(\psi,\theta; D)
\end{equation}
where $\mathcal{L}_{S}(\cdot)$ is the standard classification loss (e.g., 
cross-entropy) of the student model on $D$ and 
$\mathcal{L}_{\mathit{KD}}(\cdot)$ is a knowledge distillation (KD) loss. The 
KD loss is used to transfer the knowledge from the teacher to the student. It 
is typically defined as some form of dis-similarity between the teacher model 
and the student model. The original work~\cite{2015_hinton_distilling} on KD 
defines this similarity using the KL divergence between the outputs of the 
teacher model and the student model. This KD loss is applicable only when the 
teacher model and the student model have the same label space (i.e., the set of 
class labels in classification). In the literature, there are other KD losses 
using other information to measure this dis-similarity, including feature 
maps~\cite{romero15_iclr}, mutual relations of data 
examples~\cite{2019_Park_rkd}, attention distribution~\cite{2017_kd_attention}, 
etc. These KD losses are applicable even when the teacher model and the student 
model predict different sets of class labels. Our approach is a general 
framework and can be used together with any KD loss. 

\noindent{\bf MAML for few-shot learning.} Meta-learning is widely used for 
solving few-shot learning 
problems~\cite{2017_protonet,2017_finn_MAML,2016_matchingnet}. Our proposed 
method is built upon the model-agnostic meta-learning 
(MAML)~\cite{2017_finn_MAML} which is one of the most popular meta-learning 
approaches. MAML consists of meta-training and meta-testing. In meta-training, 
MAML learns a model $f_{\theta}$ parameterized by $\theta$ from a set of tasks. 
In few-shot classification, each training task $\mathcal{T}$ corresponds to a 
few-shot classification problem. Let $p(\mathcal{T})$ be a distribution over 
tasks and $\mathcal{T}_i \sim p(\mathcal{T})$ be a sampled task during 
meta-training. The task $\mathcal{T}_i$ has its own training set $D_i^{tr}$ 
(also known as the \emph{support set}) and validation set $D_i^{val}$ (also 
known as the \emph{query set}). The support set only contains a small number of 
training examples. Given the model parameter $\theta$, MAML obtains a 
task-adapted model parameter $\theta'_i$ by taking a few gradient updates using 
the loss on $D^{tr}_i$ 
\begin{equation} \label{eq:1}
\theta'_i \leftarrow \theta - \alpha 
\nabla_{\theta}\mathcal{L}_{\taski}\left( f_{\theta}; D_i^{tr}\right)
\end{equation}
where $\alpha$ is the learning rate. Eq.~\ref{eq:1} corresponds to the 
\emph{inner update} in MAML.

In the \emph{outer update} of MAML, we update the model parameters $\theta$ 
based on the loss of the task-adapted model on $D_i^{val}$ over training tasks
\begin{equation} \label{eq:2}
\theta \leftarrow \theta - \beta  \nabla_{\theta} \sum_{\mathcal{T}_i\sim p(\mathcal{T})}\mathcal{L}_{\taski}\left( 
f_{\theta_{i}'}; D_i^{val}\right)
\end{equation}
where $\beta$ is the learning rate. In Eq.~\ref{eq:2}, $\mathcal{L}_{\taski}$ 
is based on the task-adapted model $\theta'_i$, but the gradient update is 
performed on the model parameter $\theta$. 

The essence of MAML is to learn the initial model parameter $\theta$, so that it can effectively adapt to a new task given a small number of training examples of that task.

\section{Our Approach}
\label{sec:our approach}

\begin{figure*}[ht]
	\begin{center}
	\includegraphics[width= \textwidth]{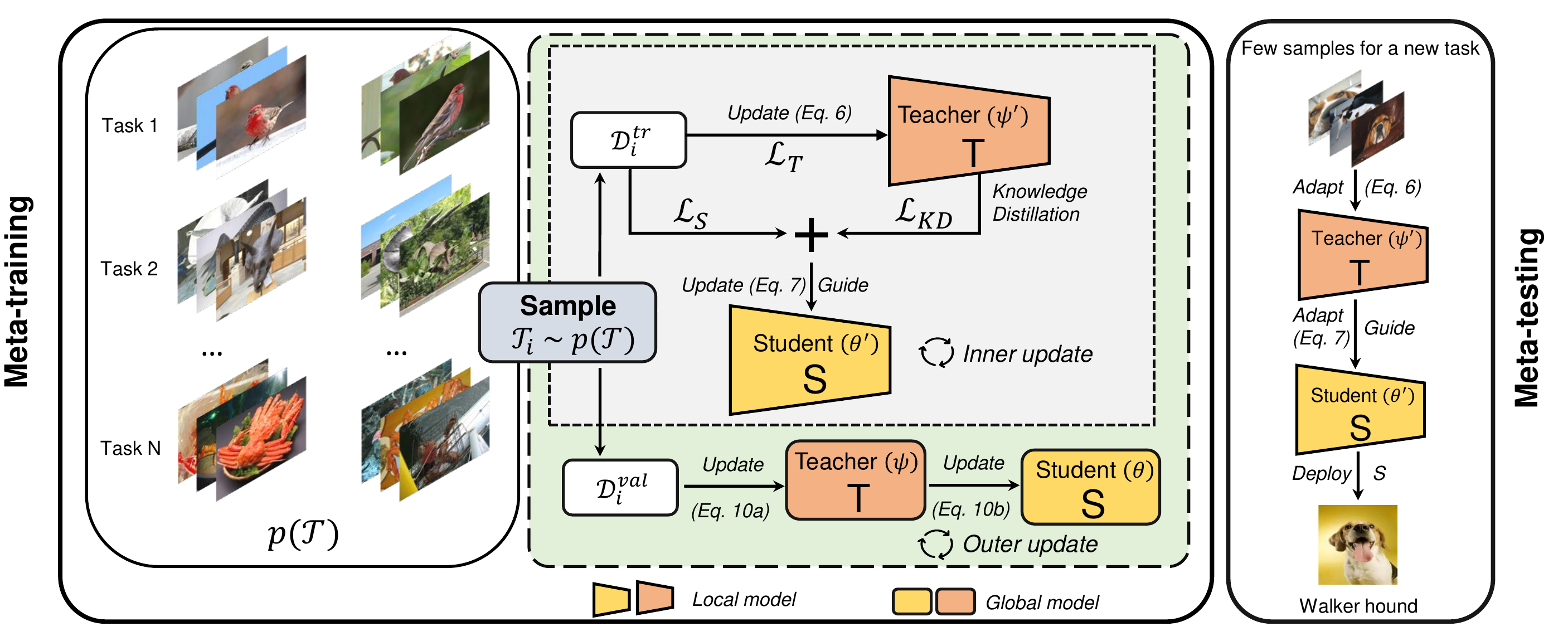}
	\end{center}
	\caption{{\bf Overview of our approach}. Similar to 
	MAML~\cite{2017_finn_MAML}, our approach has a meta-training stage and a 
	meta-testing stage. During meta-training, we jointly learn a teacher model 
	$\psi$ and a student model $\theta$. In each iteration of meta-training, we 
	sample a batch of tasks, where each task $\mathcal{T}_i$ is a few-shot 
	classification problem with its own support set $\mathcal{D}_i^{tr}$ and 
	query set $\mathcal{D}_i^{val}$. In the inner update of meta-training, we 
	obtain an updated task-specific teacher $\psi_i'$ via gradient update on 
	$\mathcal{D}_i^{tr}$. We then use $\psi_i'$ to guide the adaptation of the 
	task-specific student $\theta_i'$. In the outer loop, the global models 
	$\psi$ and $\theta$ are updated based on the performance of the 
	task-specific models on $\mathcal{D}_i^{val}$ across sampled tasks. During 
	meta-testing, we use a procedure similar to the inner update of 
	meta-training to obtain the adapted model for the new task.}
	\label{fig:overall}
\end{figure*}

In this section, we elaborate our proposed approach. An overview of our 
approach is shown in Fig.~\ref{fig:overall}.

\noindent{\bf Task-specific meta distillation.} Let $f_{\psi}(\cdot)$ be 
the teacher model parameterized by $\psi$ and $g_{\theta}(\cdot)$ be the 
student model parameterized by $\theta$. Given a task $\mathcal{T}_i=(D_i^{tr}, 
D_i^{val})$ where $D_i^{tr}$ is the support set and $D_i^{val}$ is the query 
set of this task, we use $\mathcal{L}_T(\psi; D_i^{tr})$ to denote the 
classification cross-entropy loss of the teacher model $f_{\psi}$ on 
$D_i^{tr}$, i.e.,
\begin{equation}
  \mathcal{L}_T(\psi; D_i^{tr}) = \sum_{(x_j,y_j)\in D_i^{tr}} \ell_{CE}(f_{\psi}(x_j),y_j)
\end{equation}
where $\ell_{CE}(\cdot)$ is the cross-entropy loss between the predicted class 
and the ground-truth. We similarly define the classification loss 
$\mathcal{L}_S(\theta; D_i^{tr})$ of the student model $g_{\theta}$ on 
$D_i^{tr}$ as:
\begin{equation}
  \mathcal{L}_S(\theta; D_i^{tr}) = \sum_{(x_j,y_j)\in D_i^{tr}} 
  \ell_{CE}(g_{\theta}(x_j),y_j)
  \vspace{-4pt}
\end{equation}

\begin{figure}[ht]
  \centering
\includegraphics[width= 0.45\textwidth]{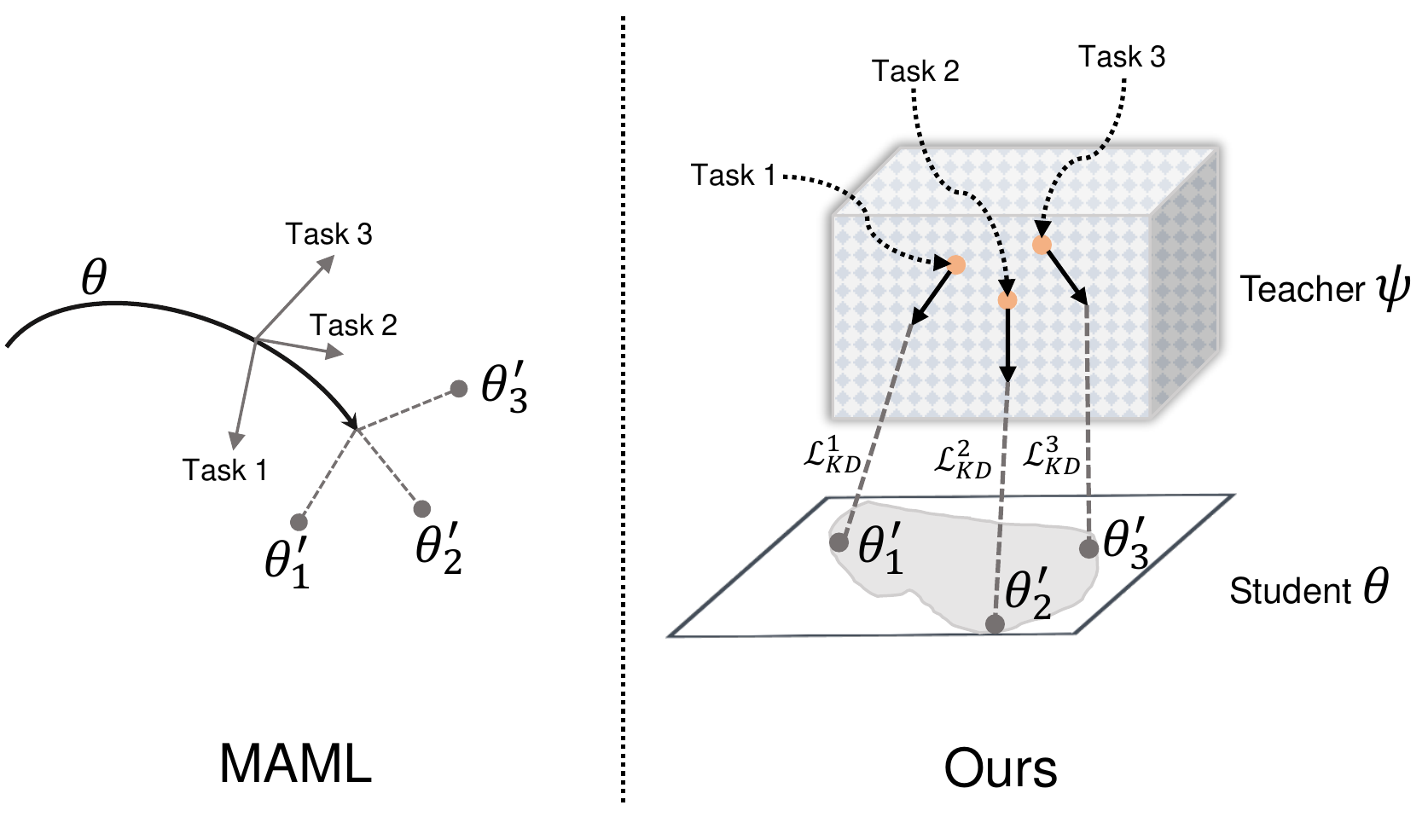}
\caption{{\bf Intuition of task-specific knowledge distillation.} In standard 
MAML, the adaptation of the model to a task only depends on the support set of 
the task. In our approach, we first adapt the teacher model using the support 
set. Since the teacher model has a higher capacity, the adapted teacher model 
can more effectively capture task-specific knowledge. The adapted teacher model 
can help guide the adaptation of the student model via the KD loss.}
\vspace{-15pt}
\label{fig:gradient}
\end{figure}

In task-specific meta distillation (TSMD), our goal is to adapt $(\psi,\theta)$ to the task $\mathcal{T}_i$ by performing a small number of gradient updates to change from $(\psi,\theta)$ to $(\psi'_i,\theta'_i)$ as follows. First we update the teacher model $\psi$ to a task-adapted model $\psi'_i$  according to the classification loss of the teacher model:
\begin{equation}
  \psi'_i \leftarrow \psi-\alpha \nabla_{\psi}  \mathcal{L}_T(\psi; D_i^{tr})
  \label{eq:inner_teacher}
\end{equation}
where $\alpha$ is the learning rate. We then update the student model $\theta$ to a task-adapted model $\theta'_i$ by transferring the knowledge from the \emph{adapted} teacher model $\psi'_i$. The motivation is that since the teacher model $\psi$ has a higher capacity, the adapted teacher model $\psi'_i$ is likely to provide useful knowledge for guiding the student model update (see Fig.~\ref{fig:gradient}). The update of the student model can be written as:
\begin{equation}
\theta'_i \leftarrow \theta-\lambda \nabla_{\theta} \Big(\mathcal{L}_S(\theta; D_i^{tr}) + \mathcal{L}_{KD}(\psi'_i, \theta; D_i^{tr})\Big)
\label{eq:inner_student}
\end{equation}
where $\lambda$ is the learning rate. In Eq.~\ref{eq:inner_student}, 
$\mathcal{L}_{KD}(\psi'_i, \theta; D_i^{tr})$ is a KD loss between the updated 
teacher model $\psi'_i$ and the student model $\theta$. Our proposed method can 
work with any well-defined KD loss.

It is worth noting the key difference between TSMD and standard KD. In standard 
KD, we assume that the teacher model is effective and we train the student 
model to emulate the teacher. In TSMD, we do not necessarily assume the teacher 
model to be effective for a new task. Instead, we assume that the teacher model 
has the capacity to successfully \emph{adapt} to a new task given a small 
amount of data. Then the student model can emulate the \emph{adapted teacher} 
model (not the original teacher model). 


\noindent{\bf Meta-training.} The updated teacher $\psi'_i$ and student $\theta'_i$ have been specifically tailored to the task $\mathcal{T}_i$. Intuitively, we would like them to perform well on the query set $D_i^{val}$ of this task. We measure their performance on $D_i^{val}$ as:
\begin{subeqnarray}
  \label{eq:loss_outer}
  &&\hspace{-15pt}\mathcal{L}_T(\psi'_i; D_i^{val})=\sum_{(x_j,y_j)\in D_i^{val}} \ell_{CE}(f_{\psi'_i}(x_j),y_j)\\
  &&\hspace{-15pt}\mathcal{L}_S(\theta'_i; D_i^{val}) = \sum_{(x_j,y_j)\in D_i^{val}} \ell_{CE}(g_{\theta'_i}(x_j),y_j)
\end{subeqnarray}

The goal of meta-learning is to learn the initial models $(\psi, \theta)$, so that after model update using the support set (Eq.~\ref{eq:inner_teacher} and Eq.~\ref{eq:inner_student}) of a particular task, the task-adapted models $(\psi'_i,\theta'_i)$ will minimize the loss defined in Eq.~\ref{eq:loss_outer} on the corresponding query set across all tasks. The meta-objective can be defined as:
\begin{equation}
  \label{eq:meta_obj}
  \min_{\psi}\sum_{\mathcal{T}_i\sim p(\mathcal{T})} \mathcal{L}_T(\psi'_i; D_i^{val}), \quad \min_{\theta}\sum_{\mathcal{T}_i\sim p(\mathcal{T})} \mathcal{L}_S(\theta'_i; D_i^{val})
\end{equation}
The meta-objective in Eq.~\ref{eq:meta_obj} involves summing over all 
meta-training tasks. In practice, we sample a mini-batch of tasks in each 
iteration during meta-training. Note that we do not need any KD loss in the 
meta-objective in Eq.~\ref{eq:meta_obj}. The KD loss is only used in the inner 
update of meta-training.

The initial models $(\psi, \theta)$ are learned by optimizing the meta-objective (Eq.~\ref{eq:meta_obj}) using stochastic gradient descent as:
\begin{subeqnarray}
  \label{eq:meta_update}
  \slabel{eq:meta_update_teacher}
  \psi \leftarrow \psi - \beta \nabla_{\psi}\sum_{\mathcal{T}_i\sim p(\mathcal{T})} \mathcal{L}_T(\psi'_i; D_i^{val})\\
  \slabel{eq:meta_update_student}
  \theta \leftarrow \theta - \eta \nabla_{\theta}\sum_{\mathcal{T}_i\sim p(\mathcal{T})} \mathcal{L}_S(\theta'_i; D_i^{val})
\end{subeqnarray}
where $\beta$ and $\eta$ are learning rates. Note that the losses 
($\mathcal{L}_T(\cdot)$ and $\mathcal{L}_S(\cdot)$) in Eq.~\ref{eq:meta_update} 
are computed using the adapted models $(\psi'_i,\theta'_i)$, but the SGD 
updates are performed over the model parameters $(\psi,\theta)$.

The high-level intuition of meta-training is to learn the teacher model $\psi$ 
and the student model $\theta$ jointly, so that the teacher model can 
effectively adapt to a new task, and the student can effectively adapt to the 
same new task using the distilled knowledge of the task-adapted teacher model. 
Algorithm~\ref{alg:tgkd} summarizes the meta-training process.

\noindent{\bf Meta-testing.} After meta-training, we obtain the model 
parameters $(\psi,\theta)$. During meta-testing, we have a new task 
$\mathcal{T}=(D^{tr}, D^{val})$ with the support set $D^{tr}$ and the query set 
$D^{val}$. We simply use Eq.~\ref{eq:inner_teacher} and 
Eq.~\ref{eq:inner_student} to obtain the task-specific parameters 
$(\psi',\theta')$. We then use the updated student model $\theta'$ for 
predictions on $D^{val}$. Note that we do not use the updated teacher model 
$\psi'$ during inference, since our assumption is that the teacher model is too 
large to be deployed for the end application.

One might argue that since we do not use the adapted teacher model $\psi'$ for the final inference, it is perhaps not necessary to update the teacher model in the meta update (Eq.~\ref{eq:meta_update_teacher}). But without Eq.~\ref{eq:meta_update_teacher}, the teacher model will stay the same throughout meta-training. This is clearly not desirable, especially if the teacher model is randomly initialized. In practice, we have found that it is crucial to update the teacher model using Eq.~\ref{eq:meta_update_teacher} to make sure that the teacher model is improved during meta-training.

\noindent{\bf Remarks.}  So far, we have assumed that the teacher model is 
updated during meta-training and meta-testing. In some applications, it might 
be difficult to update the teacher model. This can happen if the teacher model 
is a very large pretrained model (e.g., BEiT~\cite{2021_beit}, 
GPT-3~\cite{2020_brown_language}). These large models require significant 
resources and expertise to train. Once they are trained, it might be difficult 
to perform any update on these models. In some cases, the pretrained model is 
only provided as a callable API and the model itself is not directly 
accessible. Clearly we cannot update the pretrained model in these cases. In 
the meantime, these pretrained large models contain a significant amount of 
useful knowledge, so it is desirable to use them as fixed teacher models (even 
if these models cannot be updated) in our proposed framework. To handle this 
case, we simply need to slightly modify our method by omitting the update of 
the teacher model in Eq.~\ref{eq:inner_teacher} and 
Eq.~\ref{eq:meta_update_teacher}.

\begin{algorithm}[t]
	\caption{Meta-Training with Task-Specific Knowledge Distillation}
	\label{alg:tgkd}
	\begin{algorithmic}[1]
                \REQUIRE Distribution $p(\mathcal{T})$ over tasks $\mathcal{T}$ 
		\REQUIRE Learning rates $\alpha$, $\beta$, $\lambda$ and $\eta$
		\STATE Initialize teacher model $\psi$ and student model $\theta$ 
		\WHILE{not done}
		\STATE Sample a batch of tasks $\task_i \sim p(\task)$
		\FORALL{$\task_i$}
                \STATE Let $D_i^{tr}$ be the support set of $\task_i$ and $D_i^{val}$ be the query set of $\task_i$
                \STATE Obtain adapted teacher $\psi'_i$ via Eq.~\ref{eq:inner_teacher} on $D_i^{tr}$
                \STATE Obtain adapted student $\theta'_i$ via Eq.~\ref{eq:inner_student} on $D_i^{tr}$
                \STATE Evaluate meta-objective in Eq.~\ref{eq:loss_outer} on $D_i^{val}$
		\ENDFOR
                \STATE Update model parameters $(\psi,\theta)$ via Eq.~\ref{eq:meta_update}
		\ENDWHILE
	\end{algorithmic}
\end{algorithm}


\section{Experiments}
\label{sec:experiments}
In this section, we evaluate the proposed approach in MAML-based few-shot image classification on several benchmark datasets.

\subsection{Datasets and Implementation}
\noindent{\bf Datasets.} We evaluate our approach on the following standard 
benchmark datasets commonly used in few-shot image classification.

\begin{itemize}[leftmargin=*]
\item \emph{mini-ImageNet}~\cite{2016_matchingnet}: This is a standard
benchmark for few-shot image classification. It consists of
100 randomly chosen classes from ILSVRC-2012~\cite{2015_feifei_imagenet}. 
These classes are randomly split into $64$, $16$ and $20$ classes for
meta-training, meta-validation and meta-testing, respectively.
Each class contains $600$ images of size $84\times84$. 

\item \emph{FC100}~\cite{fc100}: This dataset is derived
from CIFAR-100~\cite{cifar-100}. It contains 100 classes which are
grouped into 20 superclasses. These classes are split into $60$
classes for meta-training, $20$ classes for meta-validation, and $20$ classes 
for meta-testing. Each class contains 600 images.

\item \emph{CIFAR-FS}~\cite{2018_Luca_metalearning}: This dataset contains 100 
classes from CIFAR-100~\cite{cifar-100}. The classes are randomly split
into $64$, $16$ and $20$ for meta-training, meta-validation
and meta-testing, respectively. Each class contains $600$ images.

\item \emph{FGVC-aircraft}~\cite{2013_aircraft}: This dataset contains 10,200 
images of aircraft with 100 images for each of 102 different aircraft classes. 
The classes are randomly split into $50$, $25$ and $25$ for meta-training, 
meta-validation and meta-testing, respectively. We resize images to 
$84\times84$.

\item \emph{CUB200}~\cite{2018_cub200}: This dataset contains 200 classes and 
11,788 images in total. Following the evaluation protocol~\cite{2018_cub200}, 
we randomly split the dataset into $100$, $50$ and $50$ for 
meta-training, meta-validation and meta-testing, respectively. We 
resize images to $84\times84$. 

\item \emph{Stanford dogs}~\cite{2011_dogs}: This dataset contains images of 
120 breeds of dogs. This dataset has been built using images and annotations 
from ImageNet for the task of fine-grained image categorization.  we randomly 
split the dataset into $60$, $30$ and $30$ for meta-training, meta-validation 
and meta-testing, respectively. We also resize images to $84\times84$.
\end{itemize}

\noindent{\bf Implementation details.} We use 
ResNet-50~\cite{2016_Kaiming_resnet} as the teacher network and a simple 
four-layer ConvNet (Conv-4) defined in \cite{2017_finn_MAML} as the student 
network. For mini-ImageNet, FC100, and CIFAR-FS datasets, we first 
utilize a meta-training set to train the teacher with 160 epochs for each 
dataset. Then we use the parameters to initialize the teacher model $\psi$. For 
FGVC-aircraft, CUB200, and Stanford dogs datasets, we use ResNet-50 as the 
teacher which is initialized with ImageNet~\cite{2015_feifei_imagenet} 
pretrained weights. All methods use the Adam~\cite{2015_kingma_adam} optimizer 
with initial learning rate $\texttt{1e-5}$ and $\texttt{1e-3}$ 
for teacher and student, respectively. We apply standard data augmentation 
including random crop, random flip, and color jitter. We train our model for 
about 600 epochs on mini-ImageNet, and train our model with about 300 epochs 
for other datasets.
In our experiments, we use the cross-entropy loss for classification. For 
transferring knowledge, we use the relation knowledge distillation 
(RKD)~\cite{2019_Park_rkd} loss when the teacher model is fixed, and use 
KL-divergence based KD loss~\cite{2015_hinton_distilling} when the teacher 
model can be updated. We set the number of inner update in meta-training to 5. 
We will release our code upon publication.
\begin{table}[!t]
	\begin{center}

\resizebox{0.48\textwidth}{!}{
\begin{tabular}{c||cccccc} \hline
	\phantom{a}
    & \multicolumn{2}{c}{\textbf{FC100}}
    & \multicolumn{2}{c}{\textbf{CIFAR-FS}}
    & \multicolumn{2}{c}{\textbf{mini-ImageNet}}
    \\ 
    \rowcolor[gray]{.9}
   Method & 1-shot & 5-shot & 1-shot & 5-shot  & 1-shot   & 5-shot\\
  
    \cline{1-7}
    	
    Student  & 35.97$\pm$1.80& 48.68$\pm$0.91 & 56.10$\pm$1.80 & 72.10$\pm$0.96  & 49.53$\pm$1.90 & 62.61$\pm$0.91  
    \\ 
    Fixed teacher  & 37.10$\pm$1.61& 49.19$\pm$0.92 & 56.86$\pm$1.82 & 
    72.90$\pm$0.85  & 51.27$\pm$2.02  & 63.45$\pm$0.91 
    \\ 
    \textbf{Ours}  & \textbf{38.33 $\pm$1.78} & \textbf{50.71$\pm$1.05} & 
    \textbf{57.50$\pm$1.73} & \textbf{73.23$\pm$0.93}  & 
    \textbf{51.45$\pm$1.80}  
    & \textbf{63.80$\pm$0.85} \\ \hline
    Oracle  & 44.61$\pm$1.01 & 59.60$\pm$0.70 & 72.32$\pm$0.99 & 
    85.80$\pm$0.77  & 55.60$\pm$0.96 & 66.80$\pm$0.64 \\ 
    \hline \hline
    
    \phantom{a}
    & \multicolumn{2}{c}{\textbf{FGVC-aircraft}}
    & \multicolumn{2}{c}{\textbf{CUB200}}
    & \multicolumn{2}{c}{\textbf{Stanford dogs}}
    \\ 
    \rowcolor[gray]{.9}
    Method & 1-shot & 5-shot & 1-shot & 5-shot  & 1-shot   &5-shot
    \\ \cline{1-7}
    Student  & 48.87$\pm$1.71 & 64.27$\pm$0.87 & 52.83$\pm$1.46 & 67.82$\pm$0.69 & 
    37.43$\pm$1.90    & 52.05$\pm$1.03 
    \\ 
    Fixed teacher  & 51.12$\pm$1.71 & 64.97$\pm$0.85 & 53.60$\pm$1.52 & 
    68.01$\pm$0.96 & 37.99$\pm$1.89  
    & 52.94$\pm$0.78  
    \\ 
    \textbf{Ours}  & \textbf{53.67$\pm$1.71} & \textbf{66.64$\pm$0.94} & 
    \textbf{56.13$\pm$1.76} & \textbf{69.22$\pm$0.97} & 
    \textbf{38.67$\pm$1.80}  
    & \textbf{53.89$\pm$0.89} 
     \\ \hline
    Oracle  & 61.10$\pm$1.80 & 88.20$\pm$0.74 & 66.30$\pm$0.99 & 
    84.53$\pm$0.52  & 58.32$\pm$1.49  & 75.30$\pm$0.69
    \\ \hline
\end{tabular}}
	\caption{{\bf Experimental results}. We evaluate 1-shot and 5-shot 
	classification on 6 benchmark datasets (mini-ImageNet, FC100, 
	CIFAR-FS, 
	FGVC-aircraft, CUB200 and Stanford dogs). On each dataset, We 
	report the mean of
	800 randomly generated test episodes as well as the 95$\%$ 
	confidence intervals.
	Using a fixed teacher in our approach gives better performance than 
	training 
	the student with MAML. Our full approach achieves the best 
	performance on all 
	datasets.}	
	\label{tab:results}
\end{center}
\vspace{-10pt}
\end{table}

\begin{table}[tb]
	\begin{center}
		
		\resizebox{0.45\textwidth}{!}{
			\begin{tabular}{lcccc}
				\hline                         

		 	     \phantom{a} & \multicolumn{2}{c}{\textbf{1-shot}} & \multicolumn{2}{c}{\textbf{5-shot}} \\
                \rowcolor[gray]{.9}
				\textbf{Dataset} & ResNet-50 & BEiT & ResNet-50 & BEiT\\
				\hline
				FC100  &  37.10$\pm$1.61 &  \textbf{37.90$\pm$1.42} & 49.19$\pm$0.92 &\textbf{50.22$\pm$0.69}\\
				CIFAS-FS  &  56.86$\pm$1.82  & \textbf{57.12$\pm$1.69} &
				\textbf{72.90$\pm$0.85}  &72.62$\pm$0.87\\
				mini-ImageNet  &  51.27$\pm$2.02  & \textbf{51.69$\pm$1.08} 
				&  63.45$\pm$0.91  &\textbf{63.80$\pm$0.81}\\
				FGVC-aircraft  &  51.12$\pm$1.71  & \textbf{51.30$\pm$2.00} 
				& 64.97$\pm$0.85 &\textbf{65.02$\pm$0.95}\\
				CUB200  &  53.60$\pm$1.52  & \textbf{54.22$\pm$1.62} & 
				68.01$\pm$0.96 &\textbf{68.40$\pm$0.65}\\
				Stanford dogs   & 37.99$\pm$1.89 & \textbf{38.79$\pm$1.76} & 
				52.94$\pm$0.78 & \textbf{53.57$\pm$0.77}\\
				\hline
			\end{tabular}
		           }
			\caption{{\bf Pretrained model as teacher}. We compare using an 
				off-the-shelf pretrained model (\emph{BEiT}) as the fixed 
				teacher 
				model 
				instead of \emph{ResNet-50} on different datasets. Using BEiT 
				gives 
				improved performance.}
			\label{tab:beit}
	\end{center}
\vspace{-20pt}
\end{table}

\begin{table*}[tb]
\begin{center}
\begin{small}
\begin{tabular}{@{}llc@{}cc@{}c@{}cc@{}c@{}cc@{}}
\hline
\toprule
& & \phantom{a} & \multicolumn{2}{c}{\textbf{mini-ImageNet}} & \phantom{ab} & \multicolumn{2}{c}{\textbf{CIFAR-FS}} & \phantom{ab} & \multicolumn{2}{c}{\textbf{FC100}} \\
\cmidrule{4-5} \cmidrule{7-8} \cmidrule{10-11}
\textbf{Model} & \textbf{backbone} && \textbf{1-shot} & \textbf{5-shot} && \textbf{1-shot} & \textbf{5-shot} && \textbf{1-shot} & \textbf{5-shot} \\
\hline
MAML~\cite{2017_finn_MAML} & 32-32-32-32 &&  49.53 $\pm$ 1.90  & 62.61 $\pm$ 0.91  && 56.10 $\pm$ 1.80 & 72.10 $\pm$ 0.96 && 35.97 $\pm$ 1.80 & 48.68 $\pm$ 0.91\\
ProtoNet~\cite{2017_protonet} & 64-64-64-64 && 49.4 $\pm$ 0.8 & 68.2$\pm$ 0.7 &&55.5 $\pm$ 0.7 & 72.0 $\pm$ 0.6 && 35.3 $\pm$ 0.6 & 48.6 $\pm$ 0.6 \\
RelationNet~\cite{2018_reation_net} & 64-96-128-256 && 49.3 $\pm$ 0.9 & 66.6$\pm$ 0.7 &&55.0 $\pm$ 1.0 & 69.3 $\pm$ 0.8 && - & - \\
MeTAL~\cite{baik2021meta} & 32-32-32-32 && \textbf{52.63 $\pm$ 0.37} & \textbf{70.52$\pm$ 0.39} && 56.85 $\pm$ 0.29 &  73.10$\pm$ 0.36 && \textbf{39.32 $\pm$0.33}  & 50.36 $\pm$ 0.30 \\
Ours (fixed teacher) & 32-32-32-32 && 51.27 $\pm$ 2.02 & 63.45$\pm$ 0.91 && 56.86 $\pm$ 1.82 &  72.90$\pm$ 0.85 && 37.10 $\pm$1.61  & 49.19 $\pm$ 0.92 \\
\textbf{Ours} & 32-32-32-32 && 51.45 $\pm$ 1.80 & 63.80$\pm$ 0.85 && \textbf{57.50 $\pm$ 1.73} &  \textbf{73.23$\pm$ 0.93} && 38.33 $\pm$1.78  & \textbf{50.71 $\pm$ 1.05} \\
\bottomrule
\hline
\end{tabular}
\caption{\textbf{Comparison  results.} We compare our approach to other state-of-the-arts on mini-ImageNet, CIFAR-FS and FC100. }
\label{tab:comparison}
\end{small}
\end{center}
\end{table*}

\begin{table}[tb]
		\begin{center}
			\small
			\resizebox{0.30\textwidth}{!}{
				\begin{tabular}{c|cc}
					\hline                         			
					\textbf{Method} & \textbf{1-shot} & 
					\textbf{5-shot} \\
					\hline
					Student  &  46.59$\pm$0.32 &  62.10$\pm$0.41 \\
					Fixed teacher  &  48.16$\pm$0.28  & 63.22$\pm$0.33 
					\\
					\textbf{Ours}  &  \textbf{49.59$\pm$0.28}  & 
					\textbf{64.35$\pm$0.39} 
					\\ \hline
					Oracle  &  55.12$\pm$0.36  & 66.30$\pm$0.44 \\
					\hline
				\end{tabular}
			}
		\caption{{\bf Experimental results}. We evaluate 1-shot and 5-shot 
			classification on mini-ImageNet with Reptile method.}
			\label{tab:reptile}
		\end{center}
		\vspace{-20pt}
	\end{table}	

\subsection{Baseline and Oracle Methods}
Since this paper addresses a new problem, there is no previous work that we can 
directly compare with. Nevertheless, we define several baseline methods and one oracle method for comparison.

\noindent\textbf{MAML with student network (Student).} In this baseline, we 
directly use MAML~\cite{2017_finn_MAML} to train the student network which uses 
a Conv-4 architecture.

\noindent \textbf{Ours with fixed ResNet-50 as teacher (Fixed teacher).} 
In this baseline, we use ResNet-50~\cite{2016_Kaiming_resnet} as 
the teacher model and use Conv-4 as the student model. For mini-ImageNet, 
FC100, and CIFAR-FS, we first train the teacher model with the meta-training 
set for each dataset. For FGVC-aircraft, CUB200, and Stanford dogs, we directly 
use the teacher model pretrained on ImageNet~\cite{2015_feifei_imagenet}. Then 
we train the student model using our approach while keeping the 
teacher model fixed. We use the relation knowledge distillation (RKD) 
loss~\cite{2019_Park_rkd} in this case.%

\noindent\textbf{Oracle}. In our work, we use 
ResNet-50~\cite{2016_Kaiming_resnet} as the teacher network. We consider an 
oracle method which uses ResNet-50 with MAML. This method provides an upper 
bound on the performance of our approach.

\subsection{Main Results and Analysis}\label{sec:main_results}
From the Table~\ref{tab:results}, we compare our approach with both baseline and oracle methods on six different 
datasets. Moreover, we compare to other state-of-the-art methods on mini-ImageNet, CIFAR-FS and FC100. The results are shown in Table~\ref{tab:comparison}.

As shown in Table~\ref{tab:results}, we can make several important observations. First of all, 
there is a significant gap between ``Student'' and ``Oracle''. Since both are trained with MAML and the only 
difference between them is the backbone, this performance gap confirms our 
hypothesis that MAML with a high capacity model can better adapt to new tasks. 
This also justifies ``Oracle'' being an upper bound on the performance. But of 
course, the model of ``Oracle'' might be too large to be deployed, e.g., when 
the application runs on edge devices. This is the key motivation behind our work.
Second, even with a fixed teacher, our approach (i.e., ``Fixed teacher'') 
outperforms ``Student''. This demonstrates the value of distilling the 
knowledge from the teacher model even when the teacher model is not updated 
during meta-training. The proposed approach (i.e. ``Ours") outperforms our approach (i.e. ``Fixed teacher"). This demonstrates the effectiveness of adapting teacher during adaptation stage.
Our final model (``Ours'') gives the best performance. This shows the benefit 
of jointly learning the teacher model and the student model together in a 
unified framework. Besides, we obtain the similar tendencies between 1-shot and 5-shot. 

Table~\ref{tab:comparison} shows the results compared with other state-of-the-art methods with respect to a standard 4-layer convolutional network. We can observe our proposed approach achieves good results compared with other methods, such as MeTAL~\cite{baik2021meta} and RelationNet~\cite{2018_reation_net}. MeTAL takes more benefits from a transductive setting that all query samples are available at once. We think sometimes this setting is not practical in real-world applications. 
%


\subsection{Ablation Studies}
We perform various ablation studies to gain further insights of our approach.

\noindent{\bf Pretrained model as teacher.} For results in 
Sec.~\ref{sec:main_results}, the teacher model is ResNet-50. An interesting 
question is whether it is possible to use some other large-scale pretrained 
model as the teacher. This is motivated by the current trend in the AI 
community to pretrain increasingly larger models (e.g., 
GPT-3~\cite{2020_brown_language}, BEiT~\cite{2021_beit}, BERT~\cite{2018_bert} 
and SEER~\cite{2021_facebook_self}) from very large-scale data via 
self-supervised learning. Due to the sheer size of these models, very few 
organizations in the world have the resources to train these models. However, 
once these models are trained, they can be very useful for downstream tasks 
since these models contain rich knowledge learned from large-scale data. An 
intriguing question is whether it is possible to use these large pretrained 
models in meta-learning. Due to the large size of these models, so far there is 
little work on using these models in meta-learning.

Our proposed approach can be easily modified to use a large off-the-shelf 
pretrained model as the teacher model. To demonstrate this, we choose a 
self-supervised model (BEiT)~\cite{2021_beit} as a fixed teacher model in our 
framework. The results are shown in Table~\ref{tab:beit}. Here we compare 
using fixed BEiT versus fixed ResNet-50 as the teacher model in our approach. 
We can see that using pretrained BEiT performs better in most cases. Due to the 
model size, it is difficult to directly use BEiT as the backbone architecture in MAML. Our approach provides an alternative effective way of using large pretrained models in MAML.

\noindent{\bf Reptile-based meta-learning.} Our proposed approach is based on MAML~\cite{2017_finn_MAML}. In the literature, there are other different variants of meta-learning, e.g. Reptile~\cite{reptile}. Table~\ref{tab:reptile} show results on mini-ImageNet with Reptile instead of MAML. Again, our approach outperforms other baselines. This shows that our approach is agnostic to the specific choice of the meta-learning method.

For more ablation studies, please refer to the supplementary document.


\section{Conclusions}
\label{sec:conclusions}

We have introduced a new problem called task-specific knowledge distillation in 
meta-learning. Unlike traditional meta-learning where the same model 
architecture is used in meta-training and meta-testing, our approach jointly 
learns a teacher model and a student model in the meta-learning framework. The 
adapted teacher model is then used to distill task-specific knowledge to guide 
the adaptation of the student model. We have demonstrated our approach in 
few-shot image classification with MAML. Our proposed method significantly 
outperforms vanilla MAML. Our proposed method is general and can be used to 
improve many meta-learning applications.

\noindent {\bf Limitations.} Since our approach needs to maintain both teacher and student models, the training is computationally more expensive than learning the student model alone with MAML. In this paper, we have only considered some simple KD methods. As future work, we would like to explore more advanced KD methods in our approach.

\noindent{\bf Acknowledgements.} This work is partly supported by the National Natural Science Foundation of China under Grant 62171269, the China
Scholarship Council under Grant 202006890081 and a grant from NSERC.



{\small
\bibliographystyle{ieee_fullname}
\bibliography{ref}
}

\end{document}